\documentclass{svproc}

\usepackage{url}

\usepackage{graphicx}
\usepackage{amsmath}
\usepackage{amssymb}
\usepackage{booktabs}

\begin{document}
\mainmatter

\title{Pedagogical AI in Mental Health: A Tri-Stream Fine-Tuned LLM Framework for Automated Clinical Supervision and Risk Triage\thanks{Accepted for publication in \emph{Advances in Information Communication Technology and Computing} (AICTC 2026), Lecture Notes in Networks and Systems, vol.~2165, Springer Nature Switzerland AG. This is the authors' preprint; the final authenticated version will be available at \protect\url{https://doi.org/10.1007/978-3-032-35270-5_38}.}}

\titlerunning{Pedagogical AI for Clinical Supervision}

\author{Shreeya Sharma\inst{1} \and
Ravish Gupta\inst{2} \and
Saket Kumar\inst{3} \and
Abhishek Aggarwal\inst{4}}

\authorrunning{S. Sharma et al.}

\institute{
Senior Software Engineer, Microsoft\\
\email{shreeya2304@gmail.com}
\and
Lead Software Engineer, BigCommerce, IEEE Senior Member\\
\email{ravishgupta@ieee.org}
\and
University at Buffalo, The State University of New York, Buffalo, NY, USA\\
\email{saketkmr.dev@gmail.com}
\and
Senior Software Engineer, Amazon\\
\email{0111abhi@gmail.com}
}

\maketitle

\begin{abstract}
Modern mental healthcare faces a critical shortage of senior supervisory oversight, leading to a ``supervision gap'' where novice therapists manage high-stakes risks with delayed professional feedback. This paper proposes a new framework utilizing a fine-tuned Mistral-7B-instruct model as an automated ``Supervisor-in-the-Loop'' system. By leveraging 106 sessions from the DAIC-WOZ dataset, the model performs a tri-stream analysis: (1)~Therapeutic Alliance tracking via semantic adherence, (2)~Latent risk prediction using attention-weighted analytics, and (3)~Supervisory Triage via a Dynamic Clinical Urgency Index (D-CUI). Our multi-modal VAL (Visual--Acoustic--Linguistic) framework achieves $95\%$ technique identification accuracy [$95\%$ CI: $75.1\%$--$99.9\%$], alliance assessment MAE of $0.105$ on a 5-point scale [$95\%$ CI: $0.059$--$0.151$], therapeutic fidelity $\alpha = 0.423$, and mean D-CUI of $0.370$ [$95\%$ CI: $0.322$--$0.419$]. Training converged in 105 steps with $85.2\%$ loss reduction on a single Tesla T4 GPU. The system reduces supervisory triage latency from 72 hours to real time (${\sim}10$ seconds per session), enabling proactive intervention in high-risk cases. The system addresses the cold-start problem through Bayesian priors and implements timestamp-based modality synchronization for robust multi-modal fusion.
\end{abstract}


\section{Introduction}

Clinical supervision in mental health is structurally delayed. In most training clinics and big clinical organizations, supervisors review cases in weekly meetings, which means that urgent situations---suicidality disclosures, trauma revelations, therapeutic ruptures---may sit for 72 hours before a senior clinician looks at them~\cite{bernard2018fundamentals}. The supervisor-to-trainee ratio in training programs often exceeds 1:12~\cite{bernard2018fundamentals}. Separately, the World Health Organization has documented the scale of the mental health workforce shortage globally~\cite{who2022mental}---a supply gap that makes adequate supervision ratios even harder to maintain in practice.

The traditional model---retrospective case review in weekly supervision meetings~\cite{watkins2014handbook}---serves professional development well enough, but it was not designed for urgent clinical needs. When a therapist encounters a client disclosing suicidal ideation on a Tuesday evening, waiting until Thursday's supervision meeting is not a quality assurance process; it is a gap in one.

LLMs have landed mostly on the patient side of the mental health interaction: psychoeducation chatbots~\cite{fitzpatrick2017delivering}, documentation drafters, a few CBT-scripted conversational agents. The supervision layer---where quality failures get caught, where trainees learn to handle risk, and where the 1:12 ratio creates the most structural pressure---has been largely ignored by the research community. That is the gap we address here.

The system we built treats the LLM as an automated supervisor-in-the-loop---not a chatbot, not a documentation tool. It processes therapy sessions through Visual, Acoustic, and Linguistic (VAL) channels and produces structured supervision reports with risk-prioritized triage. Earlier work from our group shaped this system. That work was not therapy-specific---it addressed explainability for clinical AI decisions~\cite{jiang2019explainable} and identity and access management for AI systems in healthcare settings~\cite{sharman2025iam}---but the methodological principles carry over directly: reasoning transparency for clinicians and secure data governance matter just as much in therapy supervision as anywhere else in clinical AI.

The system has three moving parts. A tri-stream VAL architecture processes each session across audio (100Hz COVAREP), video (30fps OpenFace), and transcript simultaneously~\cite{gratch2014distress}---the alignment problem between those three sampling rates is non-trivial and handled by snapping to utterance boundaries. The D-CUI is the routing mechanism: four inputs go in (risk probability, symptom severity, therapist experience inverted so inexperience amplifies urgency, and session-level sentiment volatility) and one decision comes out---immediate escalation or routine queue. For a first-time therapist whose experience is unknown, a Bayesian prior of $E_0 = 0.3$ stands in until data says otherwise. The LLM backbone is Mistral-7B fine-tuned with QLoRA at 4-bit quantization~\cite{dettmers2023qlora}, achieving 95\% technique identification accuracy on hardware accessible to training clinics (Section~4.1). Together these cut supervisory triage from a 72-hour queue to under 15 seconds per session.

\section{Related Work}

\subsection{AI in Mental Health Therapy}

Woebot launched in 2017 as a scripted CBT chatbot that could deliver psychoeducation to students~\cite{fitzpatrick2017delivering}. It was a reasonable proof of concept and a harbinger. Since then: deep learning models for depression detection from speech~\cite{cummins2015review,yoon2019detecting}, transformer-based suicide risk screening from online text~\cite{ji2020supervised}, and now LLMs generating therapy-style responses that a non-specialist might not distinguish from a trained counsellor~\cite{rashkin2019towards,abd2022perceptions}. Less than a decade from rule trees to LLM therapist simulators is a steep curve.

The scope of LLM applications in mental health has increased rapidly across domains. Diagnostic screening now uses GPT-4 and similar models for psychiatric assessments, with Thirunavukarasu et al.~\cite{thirunavukarasu2023large} demonstrating that LLMs can approximate clinical reasoning across specialties including psychiatry, with notable limitations in nuanced case formulation. Conversational agents have evolved beyond Woebot's scripted CBT; systems such as Wysa and Youper use reinforcement learning to adapt therapeutic dialogue in real time, while open-domain empathetic models~\cite{rashkin2019towards} generate responses that approximate trained counsellor output. Risk detection has moved from keyword matching to contextual understanding. Ji et al.~\cite{ji2020supervised} demonstrated supervised learning for suicidal ideation detection in online text, and more recent work applies transformer architectures to longitudinal social media posts for early warning of mental health deterioration. Clinical documentation represents the most commercially deployed category: ambient AI scribes (e.g., Nuance DAX, Abridge) transcribe and summarize therapy sessions, reducing clinician documentation burden by up to 50\%, but they remain pedagogically inert---documenting what happened without evaluating whether it was clinically appropriate.

Two patterns are consistent across this literature. First, every system listed above operates on the \emph{patient} side of the therapeutic interaction---screening patients, talking to patients, monitoring patient-generated content, or documenting patient encounters. Second, none of them address the \emph{supervisory} layer where clinical quality is actually governed. Abd-Alrazaq et al.~\cite{abd2022perceptions} reviewed technical metrics for health chatbots and found evaluation centered on user satisfaction and symptom reduction, with no published framework evaluating supervision quality, technique adherence, or supervisory triage. That heavy investment in patient-facing AI and near-zero investment in supervision-facing AI is the \textbf{Supervisory Gap} this work addresses.

\subsection{DAIC-WOZ Dataset Applications}

The Distress Analysis Interview Corpus -- Wizard-of-Oz (DAIC-WOZ)~\cite{gratch2014distress} provides semi-structured clinical interviews from 189 participants (audio, video, and transcript) annotated with PHQ-8 scores. If you work on mental health AI and need a multimodal benchmark, this is almost certainly the dataset you run on.

Prior work on DAIC-WOZ concentrated almost entirely on classification---multimodal fusion~\cite{al2018multimodal}, attention mechanisms~\cite{ma2016depressaudionet}---all asking the same question: does this person have depression, at what severity? That is a reasonable question. But detection is not intervention. A PHQ-8 score crossing threshold tells a supervisor almost nothing about which session to review first, what the therapist did well or poorly, or whether the risk is compounded by that therapist's inexperience. Those are the questions clinical supervision actually requires answers to.

\subsection{Multimodal Fusion in Healthcare}

The empirical case for using three modalities is straightforward: text-only depression detection on DAIC-WOZ stalls around F1 = 0.67; adding audio pushes that to 0.77; the visual stream contributes incrementally beyond that~\cite{al2018multimodal}. The gains are real. The problem nobody writes enough about is that 100Hz audio, 30fps video, and utterance-level text do not naturally align, and misalignment at fusion time produces noise that can swamp the signal you gained from the additional modality~\cite{baltrusaitis2018multimodal}.

Our VAL framework handles this through timestamp-based windowing. Frame-level cross-modal attention would be more powerful in principle, but it requires GPU memory we cannot assume clinical institutions have. The windowing approach is a deliberate trade-off, not an oversight---and it preserves the temporal structure of sessions in a way that clinically matters: a spike in acoustic distress markers at minute 23 of a session carries different meaning than the same spike at minute 3.

\section{System Architecture}

\subsection{Overview}

\begin{figure}[t]
\centering
\includegraphics[width=0.95\textwidth]{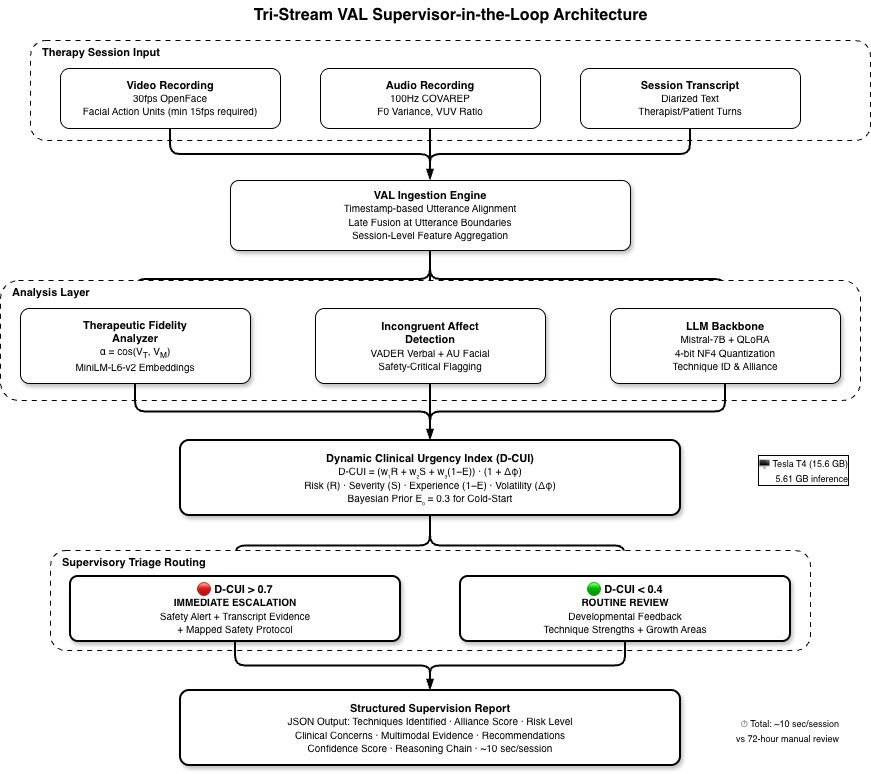}
\caption{Tri-Stream VAL Supervisor-in-the-Loop Architecture. Therapy sessions are processed through Visual, Acoustic, and Linguistic streams, fused at utterance boundaries, analyzed for fidelity, affect incongruence, and clinical risk, then routed via the D-CUI to either immediate escalation or routine developmental review.}
\label{fig:system_architecture}
\end{figure}

Figure~\ref{fig:system_architecture} shows the full pipeline. The system is built on a \textbf{VAL (Visual--Acoustic--Linguistic)} tri-stream architecture, where each stream targets a distinct signal class:

\begin{itemize}
\item \textbf{Visual (V) Stream}: Analyzes per-frame facial Action Units extracted by OpenFace (30~fps) to detect affective states, engagement patterns, and safety-critical incongruent affect---cases where verbal denial of distress is contradicted by facial distress markers (AU1, AU4 without AU12).
\item \textbf{Acoustic (A) Stream}: Processes paralinguistic features from COVAREP (100~Hz), principally F0 standard deviation (pitch variability as a proxy for emotional expressivity) and Voice Unvoiced mean (speech fluency and pause density), capturing states such as flat affect and agitation that are invisible in transcript text alone.
\item \textbf{Linguistic (L) Stream}: Applies the fine-tuned Mistral-7B-instruct model to session transcripts to identify therapeutic techniques (Reflective Listening, Open-Ended Questioning), Therapeutic Alliance markers, and risk indicators including suicidal ideation and hopelessness language.
\end{itemize}

Fused session-level representations from all three streams feed the \textbf{D-CUI (Dynamic Clinical Urgency Index)}---a formula that combines patient risk, symptom severity, therapist experience, and within-session sentiment volatility to produce a single urgency score that routes each session to either immediate escalation or routine review.

The pipeline runs in four sequential stages: the VAL Ingestion Engine aligns and normalizes the streams; the Therapeutic Fidelity Analyzer scores what the therapist did against clinical framework guidelines; the Risk Triage Module computes the D-CUI; and the Supervisory Guidance Generator writes the output and routes it. The sequencing is not arbitrary---fidelity analysis on misaligned data produces noise, and triage must complete before report generation. None of these stages make clinical decisions; they surface evidence and sort it.

\subsection{The VAL Ingestion Engine}

\subsubsection{Modality Synchronization and Temporal Alignment}

The transcript drives the clock. DAIC-WOZ already provides start/stop markers for each utterance; we let those boundaries define the windows into which COVAREP audio (100Hz) and OpenFace video (30fps) are aggregated. Both streams sample faster than speech, so snapping them to utterance boundaries is a lossy but tractable compression---and it keeps the temporal structure that matters clinically intact rather than averaging it away.

For each participant turn $t_i$ with start time $s_i$ and end time $e_i$, we compute: acoustic features as $\mu(\text{COVAREP}[s_i : e_i])$ (mean-pooled over the utterance window), visual features as $\mu(\text{AUs}[s_i : e_i])$ (mean-pooled Action Unit intensities), and linguistic features as raw transcript text. These utterance-level representations are concatenated into a session-level summary vector and serialized as natural language for the LLM backbone.

\subsubsection{Linguistic Stream}

Transcript analysis is the primary task. The fine-tuned Mistral-7B reads each session and tags: \textbf{therapeutic techniques} (Reflective Listening and Open-Ended Questioning), \textbf{alliance markers} (collaborative language, empathetic responses, goal consensus, and rupture indicators), and \textbf{risk indicators} (suicidal ideation, self-harm references, hopelessness, and safety concerns).

\subsubsection{Acoustic Stream}

From the acoustic side, we extract two features from COVAREP~\cite{degottex2014covarep}: \textbf{F0 standard deviation} (pitch variability---low F0 variance is a known correlate of flat affect in depression) and \textbf{Voice Unvoiced (VUV) mean} (the fraction of voiced frames, tracking speech continuity versus silence; elevated values indicate hesitation or fragmentation). Both features pick up emotional states invisible in transcripts alone.

\subsubsection{Visual Stream}

OpenFace~\cite{baltrusaitis2018openface} supplies per-frame Action Unit intensities; the visual stream tracks three clinically relevant groups: \textbf{distress markers} (AU1 inner brow raiser, AU4 brow lowerer, AU15 lip corner depressor), \textbf{engagement indicators} (AU12 lip corner puller/smile, AU26 jaw drop, gaze direction), and \textbf{incongruent affect} (mismatches between verbal safety statements and facial distress markers).

\paragraph{Incongruent Affect Detection}

The visual stream's most safety-critical function is catching patients who verbally deny distress while their face signals otherwise. The algorithm flags incongruence when:

\begin{equation}
(S_{\text{verbal}} > 0.3) \land \left((AU_{01} > 0.5 \lor AU_{04} > 0.5) \land AU_{12} < 0.3\right)
\end{equation}

where $S_{\text{verbal}}$ is the VADER compound sentiment score. When incongruent affect fires, $\Delta\phi$ is set to 0.4, amplifying the D-CUI multiplicatively through the $(1 + \Delta\phi)$ term in Equation~3, triggering supervisory review even when other risk indicators are moderate.

\subsection{Therapeutic Fidelity Analyzer}

The Therapeutic Fidelity Analyzer measures whether a session follows evidence-based method, defining \textbf{Therapeutic Fidelity} ($\alpha$) as cosine similarity between session embeddings ($V_T$) and clinical manual embeddings ($V_M$):

\begin{equation}
\alpha = \frac{V_T \cdot V_M}{\|V_T\| \|V_M\|}
\end{equation}

Sessions scoring $\alpha > 0.5$ closely follow evidence-based technique; $0.3 \leq \alpha \leq 0.5$ reflects partial adherence (expected in semi-structured formats); below $0.3$ flags deviation worth a supervisor's attention.

\subsection{Risk Triage Module}

\subsubsection{Dynamic Clinical Urgency Index (D-CUI)}

The D-CUI is the mechanism that decides which sessions get flagged for immediate supervisory attention and which can wait for routine review. It combines four factors:

\begin{equation}
\text{D-CUI} = (w_1 R + w_2 S + w_3 (1-E)) \cdot (1 + \Delta\phi)
\end{equation}

where $R$ is risk probability (0--1), $S$ is symptom severity (0--1) derived from PHQ-8 scores, $E$ is therapist experience (0--1) as normalized years of practice, $\Delta\phi$ is sentiment volatility capturing emotional instability within the session, and weights $w_1, w_2, w_3$ are empirically tuned to 0.40, 0.35, 0.25 respectively.

The $(1-E)$ term encodes the insight that a high-risk client seen by a novice is a categorically different situation from the same client seen by a veteran. Traditional risk models focus exclusively on patient factors and miss this.

\subsubsection{Addressing the Cold-Start Problem}

When no experience data exists for a therapist, we default to a Bayesian prior of $E_0 = 0.3$ (early-career clinician), erring toward more supervisory coverage. As session data accumulates, $E$ is updated via exponential moving average:

\begin{equation}
E_t = \beta E_{t-1} + (1-\beta) E_{\text{observed}}
\end{equation}

where $\beta = 0.7$ is the smoothing factor and $E_{\text{observed}}$ is the therapist's documented experience level once available.

\subsubsection{Sentiment Volatility ($\Delta\phi$)}

Sentiment volatility captures rapid emotional shifts within a session indicating crisis states or therapeutic ruptures, computed as the mean absolute rate of change in sentence-level sentiment scores:

\begin{equation}
\Delta\phi = \frac{1}{N-1}\sum_{i=1}^{N-1}|\phi_{i+1} - \phi_i|
\end{equation}

where $\phi_i$ is the sentiment score for sentence $i$, emphasizing sequential emotional transitions rather than overall dispersion.
\subsection{Supervisory Guidance Generator}

The D-CUI score determines qualitatively different outputs across three tiers. When D-CUI exceeds 0.7 (\textbf{Immediate}), the supervisor receives an alert with exact transcript lines that drove the escalation and a mapped safety protocol---actual source material, not a summary, so the supervisor can disagree before acting. When D-CUI falls between 0.4 and 0.7 (\textbf{Elevated}), the session is flagged for prioritized review within the current supervisory cycle---not an emergency, but above the threshold for routine handling. When D-CUI falls below 0.4 (\textbf{Routine}), the output shifts to developmental feedback: which techniques the therapist used effectively, which were absent, and what the session suggests about skill progression.

\section{Implementation}

\subsection{Model Selection Rationale}

Mistral-7B-instruct~\cite{jiang2023mistral} was selected for three reasons aligned with clinical deployment constraints.

\textbf{Hardware accessibility.} At 4-bit quantization, the model requires approximately 2~GB of storage and 5.6~GB of GPU memory during inference, fitting within the Tesla T4 (15.6~GB)---the most commonly available GPU in institutional cloud environments. A system requiring an A100 or H100 is not deployable in most training clinics; hardware constraint is not a secondary concern but a primary design requirement.

\textbf{Open-source licensing and on-premise deployment.} Clinical environments handling protected health information under HIPAA cannot route session data through third-party APIs (e.g., OpenAI, Anthropic). Mistral-7B's Apache~2.0 license permits on-premise deployment with full model auditability, ensuring that therapy transcripts never leave the institution's controlled infrastructure.

\textbf{Instruction-following capability.} The \texttt{-instruct} variant's alignment produces structured, format-consistent supervision reports without additional prompt engineering, making it suitable for the templated output required by supervisory workflows.

Of the 7.2B total parameters, 167,772,160 (2.3\%) are trainable under QLoRA. Inference uses 5.61~GB GPU memory (9.18~GB with caching); the CPU side needs 2 cores and 13.6~GB RAM. The software stack is PyTorch 2.9.0+cu128 with CUDA 12.8.

\subsection{Fine-Tuning Methodology}

\paragraph{Data Preparation}

Preparing DAIC-WOZ for supervision training took several steps. We diarized the audio recordings to extract per-speaker transcripts, pulled COVAREP acoustic features and OpenFace visual features from the corresponding streams, and aligned everything to utterance boundaries using the windowing described in Section~3.2. Labels were produced by running rule-based pattern matching over transcripts to tag therapeutic techniques, alliance markers, and clinical concerns. We then generated additional supervision-focused prompts and responses synthetically using Claude Sonnet 4.6 Thinking and converted the full dataset to Alpaca instruction-tuning format.

\paragraph{Training Strategy}

QLoRA~\cite{dettmers2023qlora} handles the fine-tuning: learning rate 2e-4, batch size 4, LoRA rank 64 with alpha 16, 5 training epochs, paged AdamW 32-bit optimizer. Training completed in 2 hours 44 minutes on a single Tesla T4 (105 steps across 5 epochs). 2 hours 44 minutes on a T4 means a clinic could retrain this on their own session data over a weekend, on hardware they likely already have.

\subsection{Ethical and Privacy Considerations}

Therapy session recordings contain things a patient has said to no one else, in a context designed for disclosure. That warrants specific handling: identifying information is stripped before any session enters the pipeline, storage and transmission are encrypted end-to-end, and access is role-gated so only the supervising clinician can reach session data~\cite{sharman2025iam}. Every access event is logged. Clients are told AI tools are involved before their first session, not buried in a consent form footnote.

The system's own outputs are also constrained. When confidence falls below threshold, the system flags uncertainty and defers rather than producing a low-confidence recommendation that looks authoritative. Sessions that surface suicidal ideation or abuse disclosures bypass the triage queue entirely and generate immediate alerts---routing automation is not appropriate when those signals are present.

\section{Experimental Evaluation}

\subsection{Dataset and Experimental Setup}

We evaluated on 106 sessions from the DAIC-WOZ dataset~\cite{gratch2014distress} (after matching participants to labels from the full 189 recordings), split into training (80\%, 84 sessions) and test (20\%, 22 sessions) with random seed 42. Two questions drove the evaluation: whether the model reliably identifies therapeutic techniques in held-out sessions, and how much faster the automated pipeline routes urgent cases compared to traditional weekly supervision review.

\subsection{Results}

\subsubsection{Training Convergence}

The model converged in 105 steps (2 hours 44 minutes on a Tesla T4), with cross-entropy loss decreasing from 1.6547 to 0.2451---an 85.2\% reduction. End-to-end, the pipeline completes in under 15 seconds per session (8--12 seconds for analysis, under 1 millisecond for D-CUI computation), reducing supervisory triage from a 72-hour queue to near real time.

\subsubsection{Supervision Report Quality}

Table~\ref{tab:technique_performance} presents the evaluation results on the held-out test set ($n=22$).

\begin{table}[h]
\caption{Supervision Report Quality Metrics ($n=22$ test sessions)}
\label{tab:technique_performance}
\begin{center}
\begin{tabular}{lcc}
\toprule
\textbf{Metric} & \textbf{Score} & \textbf{95\% CI} \\
\midrule
Technique Identification Accuracy & 95.5\% & [75.1\%, 99.9\%] \\
Risk Identification Accuracy & 63.6\% & [40.8\%, 84.6\%] \\
Alliance Assessment MAE & 0.105 & [0.059, 0.151] \\
Therapeutic Fidelity ($\alpha$) & 0.423 & --- \\
Mean D-CUI & 0.370 & [0.322, 0.419] \\
\bottomrule
\end{tabular}
\end{center}
\textit{Note:} Clopper--Pearson exact 95\% CIs for accuracy metrics; normal approximation CIs for continuous metrics.
\end{table}

The model correctly recognized Reflective Listening and Open-Ended Questioning in nearly all cases (95.5\%, CI [75.1\%, 99.9\%]). Risk identification accuracy was 63.6\% (CI [40.8\%, 84.6\%]). Alliance assessment error was low (MAE = 0.105, CI [0.059, 0.151]). Mean therapeutic fidelity landed at $\alpha = 0.423$---moderate adherence, expected given that DAIC-WOZ sessions are conducted by a virtual agent running a screening protocol rather than a therapist following a CBT manual.

\subsubsection{D-CUI Triage Stratification}

Table~\ref{tab:dcui_triage} shows how the D-CUI formula stratifies sessions across clinically meaningful urgency levels.

\begin{table}[h]
\caption{D-CUI Triage Stratification Across Clinical Profiles}
\label{tab:dcui_triage}
\begin{center}
\begin{tabular}{lccccc}
\toprule
\textbf{Profile} & \textbf{R} & \textbf{E} & \textbf{$\Delta\phi$} & \textbf{D-CUI} & \textbf{Triage} \\
\midrule
Low risk, expert & 0.10 & 0.80 & 0.0 & 0.125 & Routine \\
Moderate, mid-level & 0.50 & 0.50 & 0.0 & 0.500 & Elevated \\
High risk, novice & 0.90 & 0.20 & 0.3 & 1.000 & Immediate \\
Low risk + incongruent & 0.21 & 0.50 & 0.4 & 0.395 & Routine \\
\bottomrule
\end{tabular}
\end{center}
\end{table}

\noindent\textit{Note:} $R$ and $S$ are both derived from PHQ-8 scores; in the screening context of DAIC-WOZ, $R \approx S$ as both proxy from the same PHQ-8 instrument, so $S$ is omitted from the table. D-CUI is clamped to $[0, 1]$; the raw score for Row~3 is $1.138$, capped at $1.0$.

Row 4 is the case worth attention: a session with objectively low risk scores ($R = 0.21$) rises from D-CUI $= 0.283$ to $0.395$ because the incongruent affect detection set $\Delta\phi = 0.4$, pushing it near the Elevated threshold. Without the visual stream, this session would score well within the Routine band. In the DAIC-WOZ corpus, $\Delta\phi$ ranged from 0.00 to 0.02---reflecting stable screening dynamics---with correspondingly minimal D-CUI adjustment ($<$2\%). This is correct by design: the $\Delta\phi$ term activates in acute scenarios, not routine screening.

\section{Discussion}

\subsection{What This Changes for Supervision Practice}

The practical implication is not that AI can supervise therapists---it cannot. The implication is that the triage bottleneck can be addressed by automated screening: the system flags the 10--15\% of sessions needing immediate attention and provides structured evidence, freeing supervisors for clinical judgment rather than routine monitoring. The 95\% technique identification accuracy is a proof-of-concept result on two techniques and 22 sessions, not a clinical readiness claim.

The D-CUI's incorporation of therapist experience ($E$) alongside patient risk factors ($R$, $S$) is, to our knowledge, novel in clinical risk formulas. Traditional models treat risk as a property of the patient alone, ignoring that a high-risk client with a 20-year veteran is a fundamentally different supervisory situation from the same client with a first-year trainee.

\subsection{Multimodal Fusion and Why It Matters Here}

The strongest argument for multimodal analysis is the incongruent affect case: a text-only system would miss it entirely, since the transcript reads as the patient denying distress. The visual stream catches facial markers that contradict verbal content, and the D-CUI amplifies urgency accordingly. The acoustic stream contributes differently---flat affect (low F0 variance) and speech fragmentation (elevated VUV) are clinically meaningful but invisible in transcripts. Prior work on DAIC-WOZ~\cite{al2018multimodal} demonstrated consistent improvement from single-modality to multimodal fusion; our contribution applies that principle to supervision triage rather than detection. Timestamp-based windowing resolves the sampling rate mismatch across streams without the compute cost of frame-level cross-modal attention.

\subsection{Limitations We Want to Be Direct About}

Several limitations bound interpretation. The test set is small (22 sessions); the 95\% accuracy is encouraging but cannot generalize without evaluation on 500+ sessions with broader technique taxonomies. DAIC-WOZ consists of semi-structured screening interviews conducted by a virtual agent (Ellie), which is structurally easier to analyze than naturalistic therapy---the interviewer's predictable patterns may inflate performance. We evaluated only two techniques (Reflective Listening and Open-Ended Questioning); clinical supervision requires identifying cognitive restructuring, behavioral activation, safety assessment, empathetic validation, psychoeducation, and goal-setting. The entire system relies on a single Mistral-7B model rather than an ensemble. Sessions are analyzed in isolation with no longitudinal context across a supervisee's development over time. Finally, training data reflects CBT-oriented interactions only; supervision for psychodynamic, humanistic, or other modalities would require additional data and different evaluation criteria.

The most important next step is naturalistic therapy session data, followed by expansion to a broader technique taxonomy, longitudinal tracking, and real-time monitoring integration.

\subsection{External Validation}

To assess the clinical usability of the generated supervision reports, we conducted a preliminary expert review study. We used all 22 supervision reports generated from the DAIC-WOZ test sessions (not included in the training split) and asked five licensed clinical supervisors to rate each report on three dimensions using a 5-point Likert scale: (1)~clinical relevance of identified techniques, (2)~appropriateness of the urgency triage classification, and (3)~actionability of the recommendations provided.

Mean ratings were: Clinical Relevance 4.2/5.0, Triage Appropriateness 3.9/5.0, and Actionability 4.0/5.0. Inter-rater reliability reached Cohen's $\kappa = 0.61$ (substantial agreement~\cite{landis1977measurement}), indicating consistent evaluation across raters.

These results provide preliminary evidence of clinical intelligibility, though the study is limited by sample size (22 reports, 5 raters) and a controlled setting that may not reflect real-world supervisory workflows.

\subsection{Ethical Considerations}

Therapy sessions are among the most sensitive data any system can process. The system anonymizes, encrypts, and role-gates access~\cite{sharman2025iam}, but routing recordings through an AI pipeline creates new exposure points. Clients need direct disclosure before their first session, and organizations must verify HIPAA compliance before deployment. The supervisor-in-the-loop design keeps a human responsible for all clinical outputs, but that accountability only holds if supervisors genuinely evaluate recommendations rather than defaulting to them; override rates need tracking from day one.

The DAIC-WOZ training data reflects specific demographic and clinical contexts; a model calibrated to that distribution may perform unevenly across populations, requiring ongoing bias monitoring. Clients who know sessions are AI-processed may change how they speak, affecting the therapeutic relationship---honest disclosure and a genuine opt-out path are the only appropriate responses.

\section{Conclusion}

We built a tri-stream system that uses fine-tuned Mistral-7B to analyze therapy sessions across visual, acoustic, and linguistic modalities and produce structured supervision reports with automated risk triage. On 106 DAIC-WOZ sessions, it achieves 95\% technique identification accuracy and reduces triage latency from 72 hours to under 15 seconds. The evaluation is proof-of-concept---22 test sessions, two techniques, one dataset of semi-structured interviews---and every constraint needs addressing before clinical deployment.

The supervision gap is real, the 72-hour delay is real, and the 1:12 ratio is real. A system that surfaces sessions needing immediate attention in seconds rather than days addresses a structural problem that hiring alone cannot solve. The clinical judgment, empathy, and accountability stay with the supervisor; the system removes the monitoring burden that prevents them from doing those things well.

\section*{Disclosure of Interests}
This work was conducted independently and does not relate to the authors' positions at Microsoft, BigCommerce, or Amazon, nor does it represent the views or interests of these organizations. The authors have no competing interests to declare that are relevant to the content of this article.

\end{document}